\documentclass{article}

\usepackage{PRIMEarxiv}

\usepackage[utf8]{inputenc} 
\usepackage[T1]{fontenc}    
\usepackage{hyperref}       
\usepackage{url}            
\usepackage{booktabs}       
\usepackage{amsfonts}       
\usepackage{nicefrac}       
\usepackage{microtype}      
\usepackage{fancyhdr}       
\usepackage{graphicx}       
\graphicspath{{media/}{figures/}}     

\usepackage{amsmath}
\usepackage{amssymb}
\usepackage{enumitem}
\usepackage{svg}
\usepackage{multirow}
\setlist{nosep, leftmargin=14pt}

\title{DiffusionQC: Artifact Detection and Quality Control in Histopathology Images via Diffusion Model
}

\author{
  Zhenzhen Wang\thanks{Equal Contributions. $^{\dagger}$ Zhenzhen Wang performed the work during the internship at Biometrics Research, Merck \& Co., Inc., Rahway, NJ, USA. $^{\ddagger}$ Corresponding Authors.} $^{, \dagger}$ \\
  Johns Hopkins University \\
  Dept. of Biomedical Engineering \\
  Baltimore, MD, USA\\
  \texttt{zwang218@jhmi.edu} \\
  \And
  Zhongliang Zhou$^{\star, \ddagger}$ \\
  Merck \& Co., Inc. \\
  Biometrics Research \\
  Rahway, NJ, USA\\
  \texttt{zhongliang.zhou@merck.com} \\
  \And
  Zhuoyu Wen$^{\star, \ddagger}$ \\
  Merck \& Co., Inc. \\
  Biometrics Research \\
  Rahway, NJ, USA\\
  \texttt{wenzy0706@gmail.com} \\
  \And
  Jeong Hwan Kook \\
  Merck \& Co., Inc. \\
  Biometrics Research \\
  Rahway, NJ, USA\\
  \texttt{jeong.hwan.kook@merck.com} \\
  \And
  John B. Wojcik \\
  Merck \& Co., Inc. \\
  Translational Medicine \\
  Rahway, NJ, USA\\
  \texttt{john.wojcik1@merck.com} \\
  \And
  John Kang \\
  Merck \& Co., Inc. \\
  Biometrics Research \\
  Rahway, NJ, USA\\
  \texttt{jia.kang@merck.com} \\
}

\begin{document}
\maketitle

\begin{abstract}
Digital pathology plays a vital role across modern medicine, offering critical insights for disease diagnosis, prognosis, and treatment. However, histopathology images often contain artifacts introduced during slide preparation and digitization. Detecting and excluding them is essential to ensure reliable downstream analysis. Traditional supervised models typically require large annotated datasets, which is resource-intensive and not generalizable to novel artifact types. To address this, we propose \textbf{DiffusionQC}, which detects artifacts as outliers among clean images using a diffusion model. It requires only a set of clean images for training rather than pixel-level artifact annotations and predefined artifact types. Furthermore, we introduce a contrastive learning module to explicitly enlarge the distribution separation between artifact and clean images, yielding an enhanced version of our method. Empirical results demonstrate superior performance to state-of-the-art and offer cross-stain generalization capacity, with significantly less data and annotations.
\end{abstract}

\keywords{Computational pathology \and Diffusion model \and Artifact Detection}

\section{Introduction}

Digital pathology has become an essential part of modern oncology. High-resolution histopathology images capture details of cells, tissues, vessels, and other relevant biological structures, providing valuable insights into cancer diagnosis, treatment planning, and prognosis evaluation \cite{Digital-Pathology}. Over recent decades, computational methods, particularly AI algorithms, have been increasingly integrated into histopathology image analysis, enabling precise biomarker quantification and personalized decision-making.

\begin{figure}[!htb]
\centerline{\includegraphics[width=0.9\linewidth]{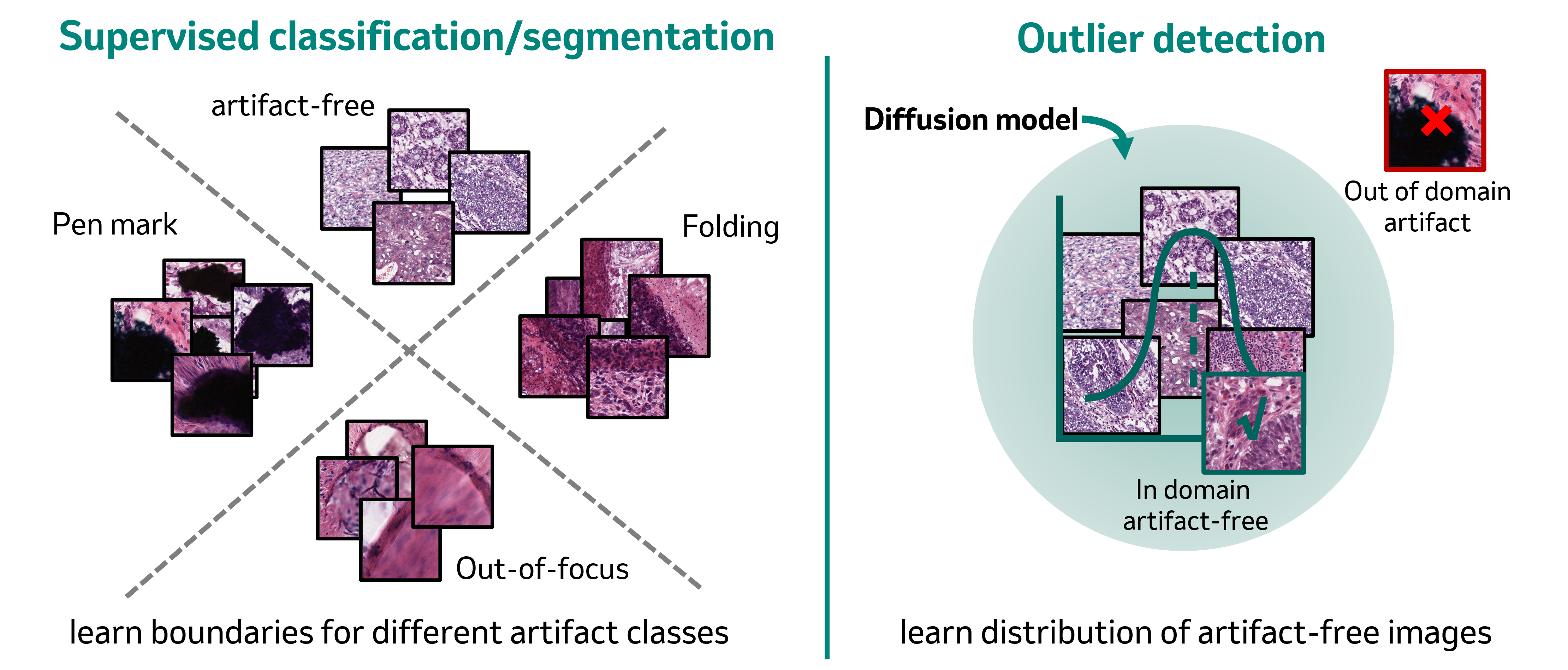}}
\caption{Conceptual comparison between supervised model and outlier detection in artifact detection.}
\label{fig:concept}
\end{figure}

Nevertheless, histopathology images frequently suffer from artifacts introduced during tissue preparation, staining, and scanning. Common artifacts include tissue folding, pen markings, pigments, air bubbles, and out-of-focus (OOF) \cite{weng2024grandqc}. They significantly impact both manual and automated analysis, particularly those relying on AI algorithms, which may fail silently on artifact samples due to morphological and textural shift. Therefore, quality control (QC) has become a critical preprocessing step in the digital pathology workflow to ensure reliability of downstream analyses. Manual delineation of artifacts is labor-intensive and impractical for large-scale datasets. Several algorithms have been developed to automate this process. Early approaches, such as HistQC \cite{janowczyk2019histoqc} and PathProfiler \cite{haghighat2021pathprofiler}, rely on handcrafted image metrics and non–deep learning algorithms, achieving only moderate accuracy. Recent deep learning–based models, including HistoROI \cite{patil2023efficient}, Kanwal et al. \cite{kanwal2024equipping}, GrandQC \cite{weng2024grandqc}, HistoART \cite{kahaki2025histoart}, and AIRAQC \cite{gautam2025airaqc}, use supervised classification or segmentation and thus require extensive manual annotations.  Among them, GrandQC curates an exceptionally large-scale set and outperforms previous methods. Although useful, these supervised models typically exhibit limited generalization to new stains or artifact types. In the real-world scenario, artifact appearance can be highly diverse, making an annotation-efficient and generalizable method urgently needed.

We propose DiffusionQC, an outlier detection-based framework for histopathology artifact identification based on latent diffusion models. It is conceptually different from the supervised methods, as we illustrate in Figure \ref{fig:concept}. Instead of classifying pre-defined artifact types, we train a model to learn the distribution of artifact-free histopathology images. During inference, samples deviating from the learned distribution are flagged as artifacts by the predicted noise residual. The proposed method has three key advantages:
\begin{itemize}
    \item \textbf{Low Annotation Needs}: Our proposed model requires only coarse delineations of artifact-free regions rather than costly pixel-wise artifact annotations.
    \item \textbf{High Performance}: Our method demonstrates comparable or superior performance to the state-of-the-art method with significantly more limited training resources. 
    \item \textbf{Promising Cross-stain Generalization}: Even in new stains with unseen artifact types, where supervised models typically fail, our model still delivered reasonable detection results.
\end{itemize}

While diffusion models have recently been explored for artifact restoration \cite{ke2023artifact, fuchs2024harp} and tumor detection \cite{linmans2024diffusion, wang2025pathology} in digital pathology, these settings assume either known artifact regions or homogeneous sample distributions. By contrast, DiffusionQC addresses a more challenging problem of detecting artifacts as out-of-distribution anomalies within a heterogeneous population of histopathology images.

\begin{figure*}[t]
    \centering
    \includegraphics[width=0.9\textwidth]{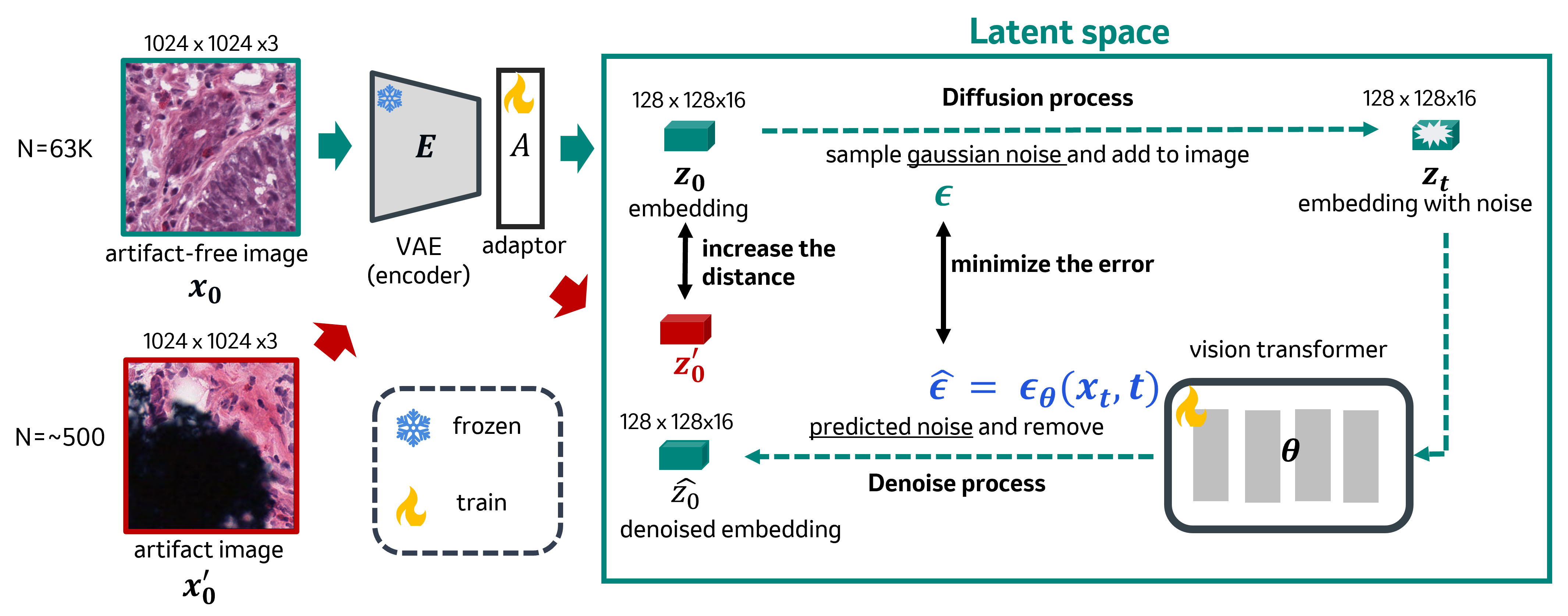}
    \caption{Illustration of DiffusionQC. Cyan arrows indicate the base framework, while red arrows denote the auxiliary contrastive learning components introduced in the enhanced version.}
    \label{fig:method}
\end{figure*} 

\section{Methods}

Rather than formulating artifact detection as a multi-class segmentation or classification problem, the proposed method, DiffusionQC, treats it as an outlier detection task. Artifact-free, clean histopathology images are regarded as in-distribution data, and artifacts are considered outliers. The goal is to learn the distribution of clean images via a latent diffusion model and to detect artifacts as out-of-distribution (i.e., outliers) during inference. We refer to this formulation as the \emph{basic} version of our method, detailed in Section \ref{sec:basic}. To further enhance the separability between images with and without artifacts, we introduce a contrastive learning component to explicitly push artifact images away from clean images in the latent space, yielding an \emph{enhanced} variant, described in Section \ref{sec:enhanced}. Figure \ref{fig:method} illustrates the overview of both variants.

\subsection{Latent diffusion model for in-distribution learning}\label{sec:basic}

Diffusion models are probabilistic generative models that learn data distribution by gradually adding noise in a \emph{forward diffusion} process and then denoising it through a \emph{reversed denoising} process, typically parameterized by deep neural networks. The latent diffusion model (LDM) is an advanced variant that first maps samplex to a latent space, where forward diffusion and reverse denoising are then conducted. We use a conditional latent diffusion transformer to learn the distribution of clean images. It consists of a variational autoencoder VAE, which maps images to the latent space, and a vision transformer (ViT), which predicts the noise added during diffusion.

Let $x_0$ be an image sampled from the clean pathology image distribution $q(x)$. The encoder of VAE, denoted as $E$, first encodes it to the latent space, giving $z_0 = E(x_0)$. The forward diffusion process forms a Markov chain that iteratively adds Gaussian noise over $T$ time steps, formulated as 
\begin{equation}
    q(z_t|z_{t-1}) = \mathcal{N}(z_t;\sqrt{1-\beta_t}z_{t-1}, \beta_tI),
\end{equation}
where $q(z_t|z_{t-1})$ denotes the distribution of a noisy embedding $z_t$ given its previous state $z_{t-1}$, and $\beta_t$ controls the noise magnitude at $t$. Since Gaussian noise addition is linear, $z_t$ can also be directly generated from the initial point $z_0$ without intermediate steps:
\begin{equation} \label{eq:forward}
    q(z_t|z_0) = \mathcal{N}(z_t;\sqrt{\bar{\alpha}_t}z_0, (1-\bar\alpha_t)I),
\end{equation}
where $\bar\alpha_t = \prod_{i=1}^t(1-\beta_t)$. This can be more intuitively written as:
\begin{equation}
    z_t = \sqrt{\bar\alpha_t}z_0 + \sqrt{1-\bar\alpha_t}\boldsymbol{\epsilon},
\label{eq:forward2}
\end{equation}
where $\boldsymbol{\epsilon} \sim \mathcal{N}(\boldsymbol{0}, I)$ represents Gaussian noise. As $t$ approaches $T$ and $1-\bar\alpha_t$ approaches 1, $z_t$ converges to a standard normal distribution. 

In the reverse denoising process, the ViT, denoted as $\theta$, predicts the noise $\boldsymbol{\epsilon}$ given the noisy embedding $z_t$ and timestep $t$, expressed as 
\begin{equation}
    \hat \epsilon = \epsilon_\theta(z_t, t).
\end{equation}
The ViT is trained to minimize the mean square error (MSE) between the actual and predicted noise with loss function:
\begin{equation}
\mathcal{L}_{basic} =  \mathbb{E}_{x,\, t,\, \boldsymbol{\epsilon}\sim \mathcal{N}(\boldsymbol{0}, I)} 
 \| \boldsymbol{\epsilon} - \epsilon_\theta(z_t, t) \|_2^2 ,
 \label{eq:loss-basic}
\end{equation}
where $x_0 \sim q(x)$ denote artifact-free training images, $t $ is uniformly sampled from $\{1, \ldots, T\}$. The noisy embedding $z_t$ is obtained by first encoding $x$ to get $z_0 = E(x_0)$ and then perturbing it as Eq.~\ref{eq:forward2}.

During inference, a query image—potentially containing artifacts—is processed identically: the VAE encodes it to the latent space, Gaussian noise is added, and the trained ViT predicts the noise. The prediction error serves as a quantification of the distribution deviation (i.e., the outlier score). Two implementation refinements are introduced during inference: 
\begin{enumerate}
    \item A constant diffusion step $t^{\star} = 800$ is set empirically and used across all images to ensure comparable error magnitudes.
    \item Pixel-wise error rather than MSE is computed to preserve spatial resolutions.  
\end{enumerate}
Concretely, an RGB image $x \in \mathbb{R}^{H_0 \times W_0 \times 3}$ is mapped by VAE to $z_0 \in \mathbb{R}^{H \times W \times C}$, where $(H, W)$ denote latent spatial dimensions and $C$ is the number of channels. The true ($\boldsymbol\epsilon$) and predicted noise ($\epsilon_\theta(z_t, t)$) share the same dimension as $z_0$. We compute an elementwise error tensor, followed by channel-wise average pooling to yield a 2D spatial error map. These errors quantify local deviations from the learned distribution, with higher values suggesting a greater likelihood of artifacts. 

\subsection{Contrastive learning to enhance distribution shift}\label{sec:enhanced}

Section \ref{sec:basic} describes the basic formulation of our method -- learning the distribution of clean images during training and detecting artifacts as outliers at inference. However, certain artifact types, such as folding or air bubbles, closely resemble clean images in appearance and morphology, making the distribution shift insufficient for a clear discrimination. To mitigate this, we introduce a lightweight adaptor parameterized by a single convolutional layer, denoted as $f_A$, appended to the VAE encoder to explicitly enlarge the distance between artifact and clean images in the latent space. This requires a small set of artifact images to guide the model. Consider a clean image $x_0 \sim q(x)$ and an artifact image $x_0^\prime \sim q^\prime(x)$, the VAE and adaptor process them sequentially to obtain latent embeddings. A contrastive loss is formulated as 
\begin{equation}
    \mathcal{L}_{con} = \mathbb{E}_{x_0,\,x^\prime_0 }\max(0, \|f_A(E(x))-f_A(E(x^\prime_0))\|_2^2 - m),
\end{equation}
where $m$ means margin, a tunable hyperparameter set as 1.2. The model is penalized when the distance between artifact and clean images falls below $m$. Incorporating the contrastive loss to the basic loss (Equation \ref{eq:loss-basic}) yields the final loss:
\begin{equation}
    \mathcal{L}_{enhanced} = \mathcal{L}_{basic} + \lambda \mathcal{L}_{con},
\end{equation}
where $\lambda$ controls the contrastive loss weights, and we set it as 0.5 empirically. By integrating the contrastive learning, we obtain an \emph{enhanced} version of the model that explicitly enforces latent space separability between clean and artifact images. At test time, a query image is processed sequentially by the VAE encoder and adaptor, and the following noise prediction and error computation are the same as the basic version.

\section{Experiments and Results}\label{sec:experiment}

\begin{table*}[t]
\centering
\caption{Quantitative results of artifact detection in H\&E-stained images.}
\begin{tabular}{l|cc|cccc|ccc}
\toprule
\multirow{2}{*}{} &
  \multicolumn{2}{c|}{\textbf{Training resources}} &
  \multicolumn{4}{c|}{\textbf{Per-type sensitivity}} &
  \multicolumn{3}{c}{\textbf{Binary artifact detection}} \\
 &
  \#WSI &
  Labels &
  \multicolumn{1}{l}{OOF} &
  \multicolumn{1}{l}{Penmark} &
  \multicolumn{1}{l}{Folding} &
  \multicolumn{1}{l|}{Air bubble} &
  \multicolumn{1}{l}{Sensitivity} &
  \multicolumn{1}{l}{Precision} &
  \multicolumn{1}{l}{F1} \\ 
  \midrule
GrandQC         & 420                 & pixel-wise                  & 0.838          & 0.970 & 0.729 & 0.884 & 0.882          & 0.671          & 0.762          \\
Ours (basic)   & \multirow{2}{*}{24} & \multirow{2}{*}{patch-wise} & 0.899          & 0.794 & 0.628 & 0.099 & 0.835          & 0.680          & 0.750          \\
Ours (enhanced) &                     &                              & \textbf{0.953} & 0.866 & 0.723 & 0.026 & \textbf{0.897} & \textbf{0.697} & \textbf{0.784} \\ 
\bottomrule
\end{tabular}
\par\vspace{2pt}
\scriptsize\textbf{Note:} The precision (and F1 score) for all methods is generally underestimated, since manual annotations omit pen marking outside tissues while algorithms typically detect them.
\label{tab:results}
\end{table*}

\begin{figure}[t]
\centerline{\includegraphics[width=\linewidth]{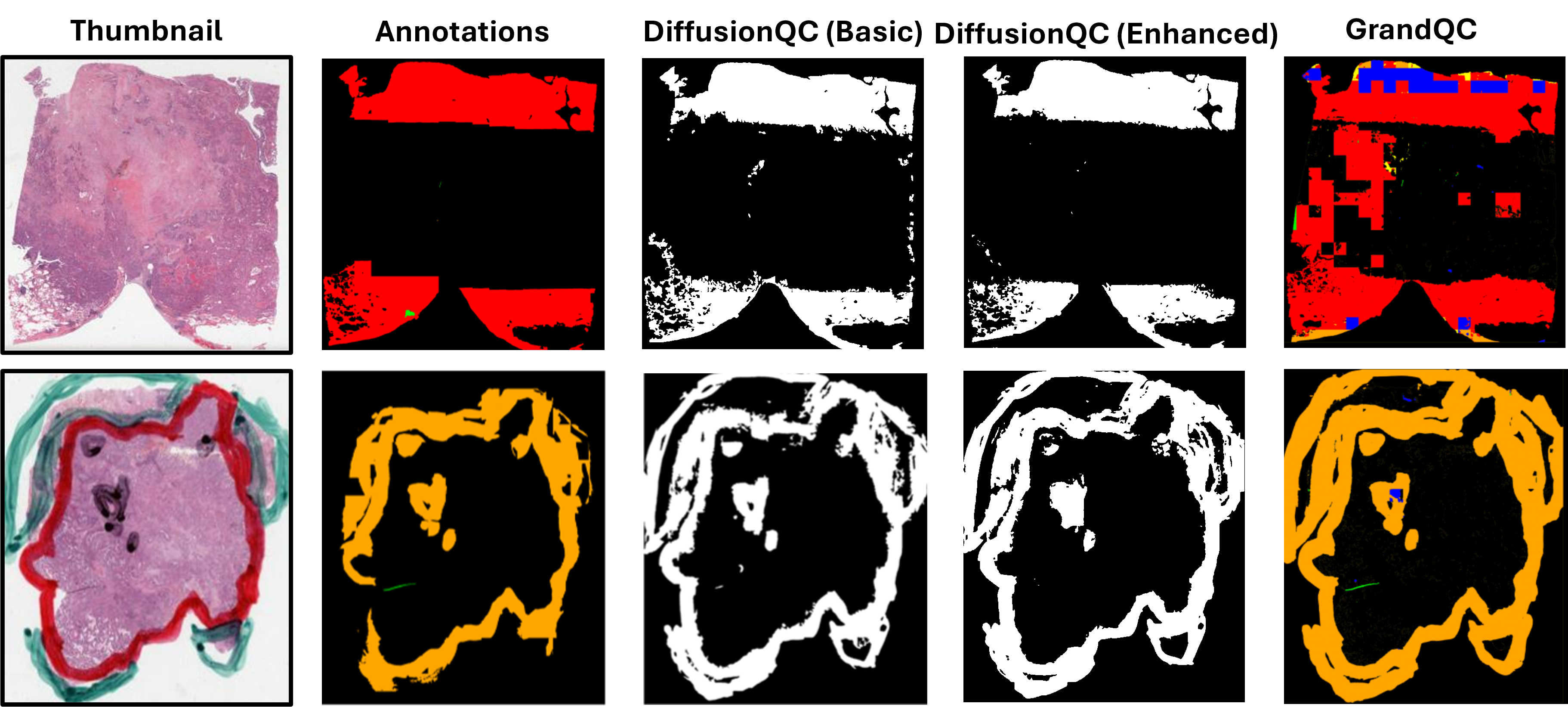}}
\caption{Artifact detection on H\&E-stained images. Columns: (1) WSI thumbnail; (2) annotations (red: OOF, orange: pen marking, lime: folding); (3–4) DiffusionQC basic and enhanced (white: artifact); (5) GrandQC (same color scheme as 2).}
\label{fig:HE}
\end{figure}

\paragraph{Model architecture:} We adopts a foundational LDM, PixCell-1024 \cite{yellapragada2025pixcell} as the base model. This model has been pretrained on a large-scale, diverse collection of histopathology images and thus captures the general image distribution across various tissues. However, artifacts were not carefully excluded during its pretraining, making the learned distribution encompass artifact appearance as well. Both the basic and enhanced versions are fine-tuned on the base model. During finetuning, the VAE remains frozen, while the ViT (and adaptor if using the enhanced version) is optimized. To make training efficient, we use Low Rank Adapted Training (LoRA) \cite{hu2022lora} for ViT fine-tuning in both versions. 

\paragraph{Dataset:} We use 50 H\&E-stained WSIs from The Cancer Genome Atlas (TCGA) spanning breast, lung, cervix, and gastric cancer. These slides include publicly available annotations for pen marking, tissue folding, out-of-focus, and air bubbles provided by AIRAQC \cite{gautam2025airaqc}. 16 WSIs are used for training, and the remaining 24 for testing. In the training set, we extract 63,000 artifact-free patches (1024 x 1024) at 10x magnification to train the basic version. Another 400 patches are sampled from artifact regions covering diverse artifact types to train the enhanced version. Notably, pixelwise annotations and artifact type labeling are not needed for training either version. 

\paragraph{Post-processing:} During inference, we use a sliding window to traverse each test WSI and extract 1024 x 1024 patches at 10x. For each patch, a spatial error heatmap is generated, with every pixel representing the noise prediction error of an 8x8 square in the original image. The raw heatmap is postprocessed with Gaussian smoothing, adaptive thresholding, and a consequential morphological close-then-open operations to generate the final binary artifact map. Thresholds are determined on a per-slide basis using an adaptive algorithm. Specifically, pixel values per slide are first restricted to $[v_{min}, v_{max}]$, and an OTSU algorithm is applied. These two user-defined hyperparameters define a confident bound: where values higher than $v_{max}$ are confidently considered artifacts and those below $v_{min}$ are considered artifact-free. Practically, they are tuned per model version and stain modality, but remain fixed across slides within the same setting.

\begin{figure}[ht]
\centerline{\includegraphics[width=\linewidth]{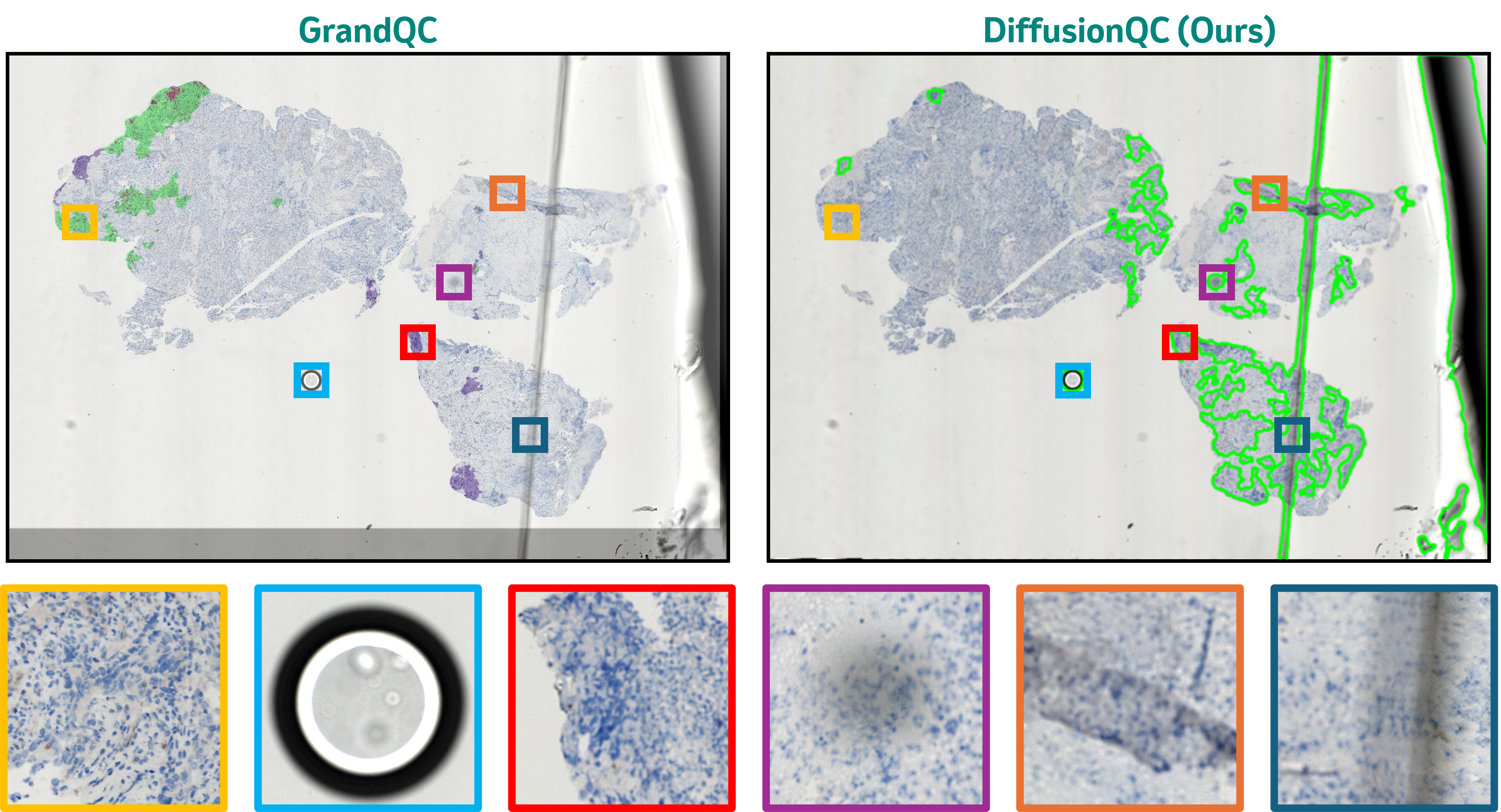}}
\caption{Cross-stain generalization on IHC-stained images. Left: GrandQC results, with artifact types overlaid (lime: dark spot; purple: out-of-focus). Right: DiffusionQC-enhanced results, with artifact boundaries marked in lime. Representative regions are boxed and magnified below.}
\label{fig:IHC}
\end{figure}

\paragraph{Results:} We evaluate the performance on 24 held-out WSIs and consider manual annotations as ground truth. For each artifact type, detection sensitivity is calculated; additionally, we aggregate all artifact categories into a binary classification task and report the sensitivity, precision, and F1 scores. We compare with the state-of-the-art method, GrandQC. As summarized in Table \ref{tab:results}, the basic version of DiffusionQC already achieves comparable performance to GrandQC, while the enhanced version outperforms it in most metrics. Importantly, DiffusionQC uses significantly fewer resources: GrandQC was trained on 420 WSIs with pixelwise annotations for artifacts, whereas ours uses only 14 WSIs with coarse patch-level labels. Several representative examples are shown in Figure~\ref{fig:HE}. Furthermore, we evaluate cross-stain generalization by directly applying the enhanced model solely trained on H\&E slides to an internal set of IHC-stained images. Although quantitative evaluation is not feasible due to the lack of ground truth annotations, we provide qualitative visualization in Figure \ref{fig:IHC}. The detection results of GrandQC and DiffusionQC are displayed side by side, with representative zoom-in patches shown below. DiffusionQC clearly demonstrates stronger detection capacity and visual consistency. Various artifacts, including air drop (blue box), OOF (red), dark spot (purple), folding (orange), and foreign objects (cyan), are successfully detected even without finetuning on IHC images. In contrast, GrandQC shows limited generalization, missing most key artifacts and falsely labeling clean regions as artifacts (yellow box).

\section{Discussion and Conclusions}

DiffusionQC reframes artifact detection and quality control as an outlier detection problem, contrasting conventional approaches that treat it as classification or segmentation. It achieves comparable or superior performance on H\&E-stained WSIs compared with the state-of-the-art, while using substantially fewer training resources. Preliminary experiments demonstrate promising cross-stain generalization capacity. Owing to the outlier detection formulation, it naturally identifies unseen or rare artifact types. The post-processing allows users to flexibly adapt thresholds to different contexts and needs. Naturally, it has some limitations. It still struggles with certain subtle artifacts, such as air bubbles. On the other hand, a trade-off of detecting less common artifacts is the risk of falsely calling uncommon tissue patterns as artifacts. Yet, these problems can be solved or relieved by carefully designing the contrastive learning penalty in the enhanced version, such as imposing heavier penalties on difficult artifact types.

\bibliographystyle{unsrt}  
\bibliography{references}

\end{document}